\documentclass{article}

\usepackage[sglblindworkshop, final]{ai4nextg_neurips_2025}

\usepackage{dsfont}

\usepackage{algorithm}
\usepackage{algpseudocode}
\usepackage{enumitem}
\usepackage{amssymb}
\usepackage{xcolor}
\usepackage{graphicx} 
\usepackage{amsmath} 
\usepackage[utf8]{inputenc} 
\usepackage[T1]{fontenc}    
\usepackage{hyperref}       
\usepackage{url}            
\usepackage{booktabs}       
\usepackage{amsfonts}       
\usepackage{nicefrac}       
\usepackage{microtype}      
\usepackage{subcaption}     

\usepackage{amsthm} 
\newtheorem{theorem}{Theorem}

\newtheorem{lemma}{Lemma}

\theoremstyle{definition}

\theoremstyle{remark}

\title{Conformal Sparsification for Bandwidth-Efficient Edge-Cloud Speculative Decoding }

\author{
Payel Bhattacharjee\thanks{Equal Contribution.
Department of Electrical and Computer Engineering; University of Arizona; Tucson, AZ, USA.
\texttt{\{payelb,fengtian,meiyuzhong\}@arizona.edu}}
\And
Fengwei Tian$^*$\\
\And
Meiyu Zhong$^*$\\
\And
Guangyi Zhang\thanks{College of Information Science and Electronic Engineering, Zhejiang University, Hangzhou 310027, China.
\texttt{zhangguangyi@zju.edu.cn}}
\And
\begin{tabular}{cc}
Osvaldo Simeone\thanks{Department of Engineering; King’s College London; London, UK;
\texttt{osvaldo.simeone@kcl.ac.uk}} &
\hspace{1.5em} 
Ravi Tandon\thanks{Department of Electrical and Computer Engineering; University of Arizona; Tucson, AZ, USA.
\texttt{tandonr@arizona.edu}}
\end{tabular}
}

\begin{document}

\maketitle

\begin{abstract}
Edge–cloud speculative decoding (SD) accelerates inference by having a cloud-based large language model (LLM) that verifies draft tokens generated by a resource-constrained small language model (SLM) at the edge. A central bottleneck is the \textit{limited bandwidth of the edge–cloud link}, which necessitates efficient compression of draft token distributions. We first derive an information-theoretic bound that decomposes the token {resampling rate} into contributions from SLM–LLM distribution mismatch and from quantization distortion. Guided by this analysis, we propose the \textit{Sparse Quantize-and-Sample SD (SQS-SD)} framework, which exploits distributional sparsity through structured sparsification and lattice-based quantization. Within this framework, \textit{$K$-SQS} applies fixed top-$K$ truncation, while \textit{C-SQS} adaptively adjusts the retained token set via online \textit{conformal prediction} to ensure bounded deviation from the dense distribution.  Empirical results confirm that both approaches improve end-to-end latency and {resampling rate} in complimentary operating regimes.
\end{abstract}

\section{Introduction}
As edge–cloud deployment of large language models (LLMs) continues to grow, efficient inference has become a critical challenge. Techniques such as model quantization and knowledge distillation \cite{polino2018model} are commonly employed to reduce model size and computational cost. However, these approaches often degrade performance by discarding fine-grained statistical information captured by larger models. A promising alternative is \textit{speculative decoding} (SD) \cite{leviathan2023fast}, where a small language model (SLM) generates multiple draft tokens that are then verified in parallel by a large LLM. This collaboration between a fast draft model and an accurate verifier boosts token throughput and lowers latency compared to standard autoregressive decoding. 


In this paper, we focus on enabling bandwidth-efficient edge–cloud SD. Specifically, we assume a SLM runs on an edge device while a more powerful LLM runs on a cloud server in tandem. A central challenge in this setup is the limited communication bandwidth between edge and cloud, which makes it imperative to compress the information (such as token probability distributions) sent over the uplink. Recent studies emphasize the importance of minimizing edge–cloud communication: for example, reference  \cite{hao2024hybrid} proposed a hybrid local–cloud inference scheme with selective offloading of tokens; the work in \cite{oh2024uncertainty} exploits uncertainty estimation to skip unnecessary uplink transmissions; and  \cite{park2025energy} explored token-level sparsification via an attention-based thresholding mechanism to reduce transmissions.  Quantization is a key component of the SD pipeline for reducing communication. In the quantize-and-sample (QS) strategy introduced in  \cite{zhang2025quantize}, the edge first quantizes the SLM’s token probability distribution and then samples from the quantized distribution, ensuring alignment with the cloud LLM’s output while also reducing the uplink payload.

QS alleviates the distribution mismatch, but communication bottlenecks can still persist due to the high dimensionality of token distributions. Importantly, prior work has shown that SLM next-token distributions are inherently sparse, with most of the probability mass concentrated in a small top-$K$ subset of the vocabulary \cite{fan2018hierarchical, hewitt2022truncation, noarov2025foundations}, while the long tail of low-probability tokens contributes negligibly. This sparsity -- often exacerbated by ``attention sinks'' in Transformer models \cite{vaswani2017attention} -- motivates the development of sparse extensions to the QS approach. Motivated by these observations, we propose \textit{sparse QS} (SQS) SD, which extends the QS framework to exploit distribution sparsity.

\begin{figure}[t]
\centering
\includegraphics[scale=0.21]{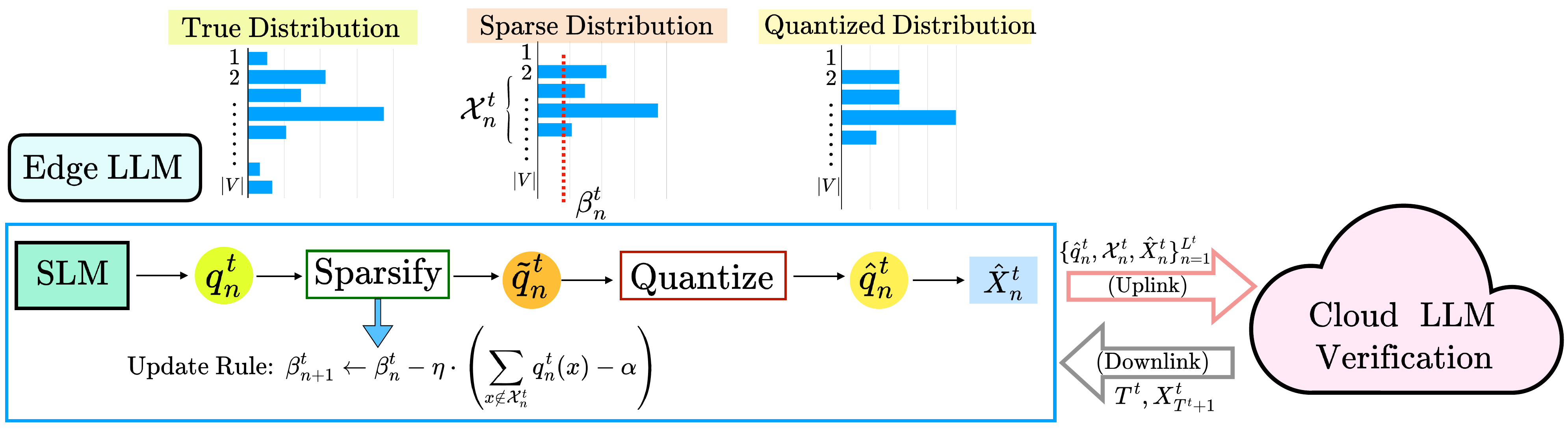}
\caption{Illustration of \textit{sparse quantize-and-sample (SQS) framework} for  edge-cloud speculative decoding for efficient LLM inference. The edge device adaptively  sparsifies and quantizes the SLM’s next-token distribution with an updated threshold with a principled online update rule based on online conformal prediction. }
\vspace{-1em}
\label{fig:workflow}
\end{figure}

\textbf{Main Contributions:} Our contributions can be summarized as follows:
\begin{itemize}[leftmargin=1.5em, itemsep=0.25em, topsep=0.25em, parsep=0pt]
\item {\textbf{Information-theoretical analysis:}} We start by  deriving an information-theoretic performance bound on the performance of SQS that decomposes the expected token rejection and resampling rate into contributions from the SLM–LLM distribution mismatch and from quantization distortion. This analysis, based on \cite{yin2024theoretical}, offers insight into the trade-offs introduced by sparsification. 

\item  \textbf{$K$-SQS:} We first present and analyze a basic SQS variant using top-$K$ truncation, which allocates bits to high-probability tokens, improving quantization resolution and reducing uplink cost.

\item \textbf{Conformal SQS (C-SQS):}  By fixing the cardinality at $K$, $K$-SQS cannot adapt to variability in next-token distributions across contexts. In practice, the SLM distribution may have widely differing effective supports, making this approach inefficient. We therefore introduce a theoretically grounded alternative, conformal QS (C-SQS), which adaptively sets the truncation threshold per token using online conformal prediction \cite{gibbs2021adaptive}. This yields a data-dependent sparsity level, guided by our theoretical bound, that ensures bounded deviation from the full (dense) QS distribution.

\item \textbf{Experimental validation:} We conduct extensive experiments demonstrating that SQS and C-SQS significantly reduce required bandwidth and end-to-end latency with negligible loss in accuracy. Our results validate the effectiveness of structured sparsification and provide quantitative guidance for navigating the latency–accuracy trade-off in edge–cloud LLM inference.
\end{itemize}

\vspace{-10pt}
\section{Sparsify-Quantize-and-Sample (SQS) Speculative Decoding}
\vspace{-10pt}
This section reviews edge-cloud speculative decoding (SD) and the quantize-and-sample (QS) method from \cite{zhang2025quantize}. We then extend QS with sparsification, introducing sparse QS (SQS) to further reduce bandwidth. We present an information-theoretic analysis of SQS performance, forming the basis for protocol design in the following sections. Finally, we introduce a simple instantiation of SQS based on top-k selection.


\textbf{Speculative Decoding:} 
Edge-cloud SD accelerates LLM inference by parallel verification of the tokens generated by a small draft model (SLM) at the edge via a larger target model (LLM) at the cloud \cite{leviathan2023fast}. SD operates in batches. At each batch $t$, the SLM takes the current prefix of accepted tokens  to generate $L^t$ new tokens in an autoregressive manner. The $n$-th generated token,  denoted as  $\hat{X}_n^t$ is sampled from the SLM distribution $ {q}_n^t$, where $V$ is the size of the vocabulary $\mathcal{V}=\{1,...,V\}$. The notation $ {q}_n^t(\cdot)$ is an abbreviation for the conditional distribution $ {q}(\cdot|C^t_n)$, where $C^t_n$ represents the context of all previously generated tokens, including all prior batches. A similar notation will be used for the LLM distribution.  

The draft token sequence $\{\hat{X}_1^t,\ldots,\hat{X}_{L^t}^t\}$ and the associated distributions $\{ {q}_1^t,\ldots, {q}_{L^t}^t\}$ are conveyed to the cloud, which uses them via the target LLM for parallel verification. Specifically, to verify the $n$-th token $\hat{X}_n^{t}$ in the batch, the target LLM computes its corresponding conditional distribution ${p}^t_n$ and accepts it if the inequality $ {q}_n^t(\hat{X}_n^t)\leq  {p}^t_n(\hat{X}_n^t)$ holds, and otherwise rejects it with probability $(1-{p}^t_n(\hat{X}_n^t)/ {q}_n^t(\hat{X}_n^t))$. Denoting  $T^t$  as the number of accepted tokens $\{{X}_n^t=\hat{X}_n^t\}_{n=1}^{T^t}$, with $T^t\leq L^t$, the cloud samples a new token ${X}_{T^t+1}^t$ from the adjusted distribution $ {\bar{p}}^t_{T^t+1}(\cdot)\propto (\max(0, {p}^t_{T^t+1}(x)- {q}^t_{T^t+1}(x)))$ if $T^t<L^t$ and from distribution $p^t_n (\cdot)$ if $T^t=L^t$. This procedure is repeated across batches $t=1,2,\ldots$ producing a sequence of $T=\sum_t (T^t+1)$ verified tokens $\{{X}_n^t\}_{n=1}^{T}$. These tokens follow the same distribution as those generated autoregressively by the cloud \cite{leviathan2023fast}.

\textbf{Quantize-and-Sample Speculative Decoding:} In edge–cloud SD with a bandwidth-limited uplink channel from edge to cloud, the edge must  compresses the edge-based SLM distribution ${q}_n^t$ into a quantized form $\hat{ {q}}_n^t$ prior to communicating them to the cloud. QS, introduced in \cite{zhang2025quantize}, generates the draft tokens $\{\hat{X}_1^t,\ldots,\hat{X}_{L^t}^t\}$ using the quantized distributions $\{\hat{q}_1^t,\ldots,\hat{q}_{L^t}^t\}$. This has the key advantage that the resulting accepted tokens $\{{X}_n^t\}_{n=1}^{T}$ preserve the same distribution as for LLM-generated tokens, thus maintaining the key guarantee of SD.

\textbf{Sparsify-Quantize-and-Sample Speculative Decoding:} While QS reduces the communication bandwidth via quantization, further gains can be achieved by exploiting the inherent sparsity of draft token distributions. In fact, prior works \cite{fan2018hierarchical,hewitt2022truncation,noarov2025foundations} have shown that next-token probabilities are typically skewed, with most of the mass concentrated in a small subset. Building on this observation, we extend QS to allow for a preliminary sparsification step. We refer to the resulting family of procedures as SQS.

Specifically, as shown in Fig \ref{fig:workflow}, the SLM probability distribution  $ {q}_n^t $ is first sparsified into a distribution $ {\tilde{q}}_n^t $ with support set  
$\mathcal{X}_n^t \subseteq \mathcal{V}$ encompassing $K^t_n \leq V$ terms, and then quantized into a distribution  $\hat{ {q}}_n^t$. By the properties of QS, the resulting protocol also guarantees that the accepted tokens have the same distribution as for the LLM in the cloud. 

We adopt with sparse lattice quantization (SLQ) \cite{teku2024latency,teku2024communicating}, which maps the retained $K^t_n$ probabilities onto a structured lattice within the probability simplex. SLQ is characterized by a resolution parameter $\ell^t_n$, with a smaller $\ell^t_n$ achieves coarser quantization at reduced bit cost. 

With any SQS protocol, the number of bits required to represent the quantized vector $\hat{ {q}}_n^t$ are given by \begin{equation} b_n^t(K_n^t,\ell_n^t) = \tilde{b}_n^t(K_n^t) + \hat{b}_n^t(K_n^t,\ell_n^t),\end{equation} where the first term, $\tilde{b}_n^t(K_n^t)$, represents the number of bits required to describe the subset $\mathcal{X}_n^t$, and the second term, $\hat{b}_n^t(K_n^t,\ell_n^t)$, is the number of bits needed to describe the non-zero elements in vector $\hat{ {q}}_n^t$. The term $\tilde{b}_n^t(K_n^t)$ depends on the specific SQS scheme, while the second term is given by  \cite{teku2024latency,teku2024communicating}\begin{equation} \hat{b}_n^t(K_n^t,\ell_n^t)=\log_2 {\ell_n^t+K_n^t-1 \choose K_n^t-1}.\end{equation}

    \paragraph{Information-Theoretic Analysis of SQS:} Building on the theoretical guarantees in \cite{yin2024theoretical}, the following result establishes an upper bound on the average number of rejected tokens $N_{\mathrm{rej}} $. As in \cite{yin2024theoretical},   the number of rejected tokens, $N_{\mathrm{rej}} $,  represents here the number of tokens that are sampled at the edge and then rejected and resampled at the cloud. Note that there is at most one rejected and resampled token per batch. Appendix \ref{app:proof_theorem1} provides further discussion on this definition. 
    
    The bound presented below  isolates two separate contributions to the average number of rejected (and resampled) tokens: 
        (\textit{i}) the intrinsic statistical mismatch between the SLM draft distribution and the LLM target distribution \cite{yin2024theoretical}, and 
        (\emph{ii}) the  distortion introduced by sparsification and quantization.  We denote as $\mathrm{TV}(p, q)$ as the total variation distance between $p$ and $q$.
    For the proof of this result (Theorem \ref{the:LASER}), we refer the readers to the Appendix \ref{app:proof_theorem1}.
    
    \begin{theorem}\label{the:LASER}
        Consider a sequence of $T$ tokens $\{X_t\}^T_{t=1}$ generated by using an SQS protocol, with corresponding per-token subsets $\mathcal{X}_n$ of cardinality $K_n(\mathcal{X}_n)$ and resolution parameters $\ell_n$ for the tokens $n=1,...,T$. The expected number of rejected tokens can be upper bounded as
        \begin{equation}
            \small
            \mathbb{E}[N_{\mathrm{rej}}]\small\leq \underbrace{\sum_{n=1}^{T}\mathbb{E}_{{\{X_t\}^{n-1}_{t=1}}\sim p}\!\left[\mathrm{TV}\!\left(q_n(\cdot\mid \{X_t\}^{n-1}_{t=1}),\,p_n(\cdot\mid \{X_t\}^{n-1}_{t=1})\right)\right]}_{\text{SLM-LLM discrepancy}}+\small\underbrace{\sum_{n=1}^{T}\left(\alpha_n(\mathcal{X}_n) + \frac{K_n(\mathcal{X}_n)}{4\ell_n}\right),}_{\text{SLQ-based distortion}}
        \end{equation} 
        where the expectation $\mathbb{E}_{\{X_t\}^{n-1}_{t=1}\sim p}[\cdot]$ is with respect to tokens generated from the LLM model,  and  the notation 
        \begin{equation} 
            \alpha_n(\mathcal{X}_n) =  \sum_{x \notin \mathcal{X}_n^t} \mathbb{E}_{\{X_t\}^{n-1}_{t=1}\sim p} [q(x \mid \{X_t\}^{n-1}_{t=1})]
        \end{equation} 
        represents  the average total probability in the subset of dropped tokens (i.e., tokens not selected during sparsification). 
    \end{theorem}

    \paragraph{Top-$K$ Sparsify-Quantize-and-Sample Speculative Decoding:} In the next we propose a novel instantiation of SQS. 
    We start here with a simple approach, referred to as \emph{$K$-SQS}, in which the subset $\mathcal{X}^t_n$ defining the support of the quantized distribution is selected by following the standard top-$K$ selection rule for a fixed value of the hyperparameter $K$. 
    Accordingly, the subset $\mathcal{X}^t_n$ encompasses the $K$ terms $x\in \mathcal{V}$ in the vocabulary with the largest probabilities $q^t_n(x)$ under the SLM. 
    For $K$-SQS, the overhead for representing the subset $\mathcal{X}^t_n$ is given by \begin{equation}\tilde{b}_n^t(K)=\log_2 {V\choose K},\end{equation} since there are ${V\choose K}$ possible subsets of dimension $K$. 
    A performance analysis for $K$-SQS follows directly from the general result in Theorem 1. 
    In particular, the SLQ-based distortion term in the upper bound  (Theorem \ref{the:LASER}) is characterized by a fixed value $K_n(\mathcal{X}_n)=K$, while  the value of the probability $\alpha_n(\mathcal{X}_n)$ varies across index $n$, as the total mass in the dropped tokens changes across token generation.
\section{Conformal Sparsify-Quantize-and-Sample Speculative Decoding}
    \label{sec:conformal}
    \textbf{Motivation for Adaptive Sparsification:} By fixing the cardinality at $K$, $K$-SQS cannot adapt to variability in next-token distributions across contexts. 
    In practice, the distribution $q^t_n$ may have widely differing effective supports across indices $t$ and $n$ \cite{jin2025massive, holtzman2019curious}, making this approach inefficient. 
    For example, after the prompt “\texttt{The capital of France is},” the continuation is highly predictable (“\texttt{Paris}”), allowing aggressive sparsification—i.e., a small cardinality $K_n^t$ for subset $\mathcal{X}^t_n$—without quality loss. 
    In contrast, a context such as “\texttt{She opened the box and found}” admits more uncertainty in the next-token prediction, requiring a larger support set $\mathcal{X}^t_n$ with larger cardinality $K_n^t$.
\newline
\newline
    \textbf{Adaptive Thresholding:} To overcome this limitation, this section introduces an adaptive thresholding strategy based on online conformal prediction \cite{gibbs2021adaptive,angelopoulos2023conformal,zecchin2024localized}. 
    In this scheme, referred to as \emph{conformal SQS} (C-SQS), the support set is determined as  
    \begin{equation}\label{eq:SQSset}
        \mathcal{{X}}^{t}_n(\beta_n^t) = \{ x \in \mathcal{V} : q_n^t(x) \ge \beta^t_n \},
    \end{equation} 
    where the sequence of thresholds  $\beta^t_n $ is designed to control the average number of rejected tokens via the upper bound (\ref{the:LASER}).

   

    Specifically, the key idea behind C-SQS is to vary the threshold $\beta^t_n $ in (\ref{eq:SQSset}) in such a way so as to ensure that the term $\sum_{n=1}^{T}\alpha_n(\mathcal{X}_n) $ in the upper bound \eqref{the:LASER} does not grow too quickly with the number of tokens $T$. 
    This design goal is motivated by the result in Theorem \ref{the:LASER}, which suggests that controlling this term provides a way to limit  the number of rejections.  
    We specifically aim at guaranteeing an upper bound of the form:
    \begin{equation}\label{eq:req}
       \frac{1}{T} \sum_{n=1}^{T} \alpha_n(\mathcal{X}_n)  \leq \alpha  + \frac{C}{T},
    \end{equation}
    where $\alpha\in(0,1)$ is a target value, $C$ is some arbitrary fixed positive constant, and index $n$ runs over the accepted tokens. The requirement (\ref{eq:req}) ensures that the term in the upper bound (\ref{the:LASER}) that is negatively affected by sparsification, namely $\sum_{n=1}^{T} \alpha_n(\mathcal{X}_n)$, is asymptotically no larger than a target value $\alpha$. 
    The choice of hyperparameter $\alpha$ dictates the degree to which we wish C-SQS to be aggressive in sparsification. 
    A larger $\alpha$ allows for a larger growth of the distortion due to sparsification, which in turn makes it possible to reduce the second term in the SLQ-based distortion in (\ref{the:LASER}). 
    The fixed value of $K$ used in $K$-SQS in contrast, carries no operational significance in terms of the final performance of the algorithm, the hyperparameter $\alpha$ thus relates directly to the protocol's performance via Theorem 1, controlling the trade-off between sparsification distortion and bandwidth reduction.

   In order to ensure (\ref{eq:req}), SQS adopts an update rule based on online conformal prediction \cite{angelopoulos2024theoretical}, whereby the threshold is updated during token generation as follows:
    \begin{equation}
        \label{eq:update}
        \beta^t_{n+1}= \beta^t_{n}-\eta \cdot \left(\sum_{x \notin \mathcal{X}_n^t} q_n^t(x)-\alpha\right),
    \end{equation}         

    where $\eta$ is the learning rate. 
    The sum $\sum_{x \notin \mathcal{X}_n^t} q_n^t(x)$ corresponds to the mass of the SLM distribution that is not contained in the support set $\mathcal{X}_n^t$. 
    When averaged over the accepted tokens, this quantity yields the term $\sum_{n=1}^{T'} \alpha_n(\mathcal{X}_n)$, where $T'$ is the total number of tokens accepted so far. 
    Intuitively,  if the probability $\sum_{x \notin \mathcal{X}_n^t} q_n^t(x)$ exceeds the target value $\alpha$, then the retained support of the distribution is too small and one needs to decrease the threshold. 
    Conversely, one can increase the threshold.

    Since the average in the upper bound (\ref{the:LASER}) is only over tokens generated from the LLM, as summarized in Algorithm 1, SQS implements a check-pointing and backtracking strategy, whereby the update (\ref{eq:update}) is first applied at the edge for all tokens for each batch. 
    Once feedback is received from the cloud, the value of the threshold is returned to the value of the last accepted token, and a further iteration is done for the new, $({T^t+1})$-th token. 
 \begin{algorithm}[h]
        \caption{Conformal Sparse Quantize-and-Sample Speculative Decoding (C-SQS)}
        \label{alg:csqs_edge_cloud}
        \begin{algorithmic}[1]
            \State \textbf{Input:} Initial context, vocabulary $\mathcal{V}$, target deviation $\alpha$, learning rate $\eta$, initial threshold $\beta^1_1$, per-batch number of tokens  $L^t$, resolution parameters $\ell^t_n$
            \For {each batch $t = 1,2,\dots$}
                \State Set $n \gets 1$ 
                \For {each token $n=1,...,L^t$}
                    \State Compute next-token SLM distribution $q^t_n(\cdot )$ 
                    \State Evaluate support ${\mathcal{X}}^t_n(\beta^t_n) $ and cardinality $|\mathcal{X}^t_n(\beta^t_n)|$ 
                    
                    \State Apply SLQ to $\tilde{q}^t_n$ to obtain quantized SLM distribution  $\hat{q}^t_n$ and sample $X^t_n \sim \hat{q}^t_n$
                    \State Apply threshold update (\ref{eq:update})
                
                \EndFor

                \State Transmit $\{ \hat{q}^t_n, {\mathcal{X}}^t_n(\beta^t_n), X^t_n \}_{l=1}^{L^t}$ to the cloud.
                \State Receive $T^t$ (number of accepted tokens) and new token $X^t_{T^t+1}$ from the cloud
                \State Apply the update (\ref{eq:update}) on the new token as $
                \beta^t_{T^t+1}= \beta^t_{T^t}-\eta \cdot \left(\sum_{x \notin \mathcal{X}_n^{T^t}} q_n^{T^t}(x)-\alpha\right)$
                \State Initialize next batch with $\beta^{t+1}_1 \gets \beta^t_{T^t+1}$ 
            \EndFor
        \end{algorithmic}
    \end{algorithm}
    \newline
    \textbf{Communication Overhead:}
    Since C-SQS varies the support $K_n^t$ along the generated tokens, the edge must communicate to the cloud both the size of the subset and the specific subset. 
    This yields the number of bits$ \tilde{b}_n^t(K^t_n)= 
            \left\lceil \log_2 \binom{|V|}{K^t_n} \right\rceil + 
            \left\lceil \log_2 |V| \right\rceil,$where the second term, $\log_2 |V| $ represents the additional overhead required to communicate the value $K_n^t$.
\newline
    \textbf{Theoretical Guarantee: }The following theorem, as proved in Appendix \ref{app:proof_theorem2}, justifies the use of the update rule (\ref{eq:update}).  
    
        
    \begin{theorem}\label{lem:csqs_bound_all} 
    For any learning rate $\eta>0$, C-SQS satisfies the requirement (\ref{eq:req}) as 
        \begin{equation}
           \frac{1}{T} \sum_{n=1}^{T} \alpha_n(\mathcal{X}_n) \leq \alpha+  \frac{|\beta^1_1|+1+\eta\alpha}{\eta T}.
        \end{equation}
    \end{theorem}




\section{Experiments and Discussion}
\textbf{Evaluation Setup and Objectives:} 
    We have conducted text completion experiments on the One Billion Word Benchmark (LM1B) dataset \cite{chelba2013one}. 
    We have used GPT-Neo-125M \cite{GPT-Neo-125m} as the edge SLM,  and GPT-Neo-1.3B \cite{GPT-Neo-1.3b} as the cloud LLM. 
    For evaluating and comparing $K$-SQS and C-SQS, we analyze two key performance metrics: (a) the average end-to-end latency (average total time),  consisting of the SLM computation time, uplink communication time, and cloud LLM verification time as detailed in \cite{zhang2025quantize}; and (b) average resampling rate, i.e., the ratio between the average number of rejected and resampled tokens, $N_{\textrm{rej}}$ and the total number of batches. 
    We vary the sampling temperature of the SLM and LLM between $[0,1]$,  while setting the quantization resolution $\ell$ and the per-batch uplink budget $B$ (in bits) as $B = 5000$ and $\ell=100$.   
     For each batch $t$, the number of generated tokens is selected as $L^t = \max\{\,L\in\mathbb{N}_0 \;:\; \sum_{n=1}^{L}  b_n^t(K_n^t,\ell) \;\le\; B \,\}$.
    This condition is enforced in a sequential way, stopping token generation when the bit budget is exhausted. 
     For C-SQS, we have used a fixed learning rate $\eta =0.001$ and deviation parameter $\alpha = 0.0005$.
    


 \begin{figure}[t]
    \centering
    \includegraphics[scale = 0.46]{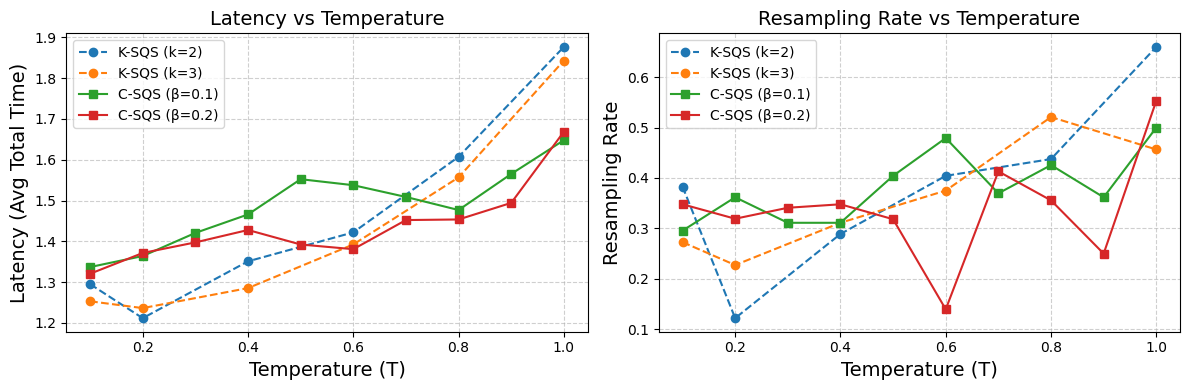}
    \caption{{Latency (average total time in seconds) and resampling rate for $K$-SQS and C-SQS across different temperatures $(T)$. $K$-SQS shows increasing latency and higher variability in resampling rate with increase in $T$, while C-SQS maintains more stable performance, achieving a better trade-off between latency and resampling efficiency in higher-uncertainty regimes.}}
    
    
    \vspace{-1em}
    \label{fig:batches}
\end{figure}
  \noindent \textbf{Results:}  {Figure~\ref{fig:batches} shows the average latency  and resampling rate for $K$-SQS and C-SQS across different temperatures $T$. At low temperatures, the draft distribution is sharply peaked, so the target token typically falls within the top-$K$ set for a fixed $K$. In this regime, $K$-SQS yields fewer rejections and resampling than C-SQS, provided $K$ is chosen appropriately. This in turn yields a lower latency for $K$-SQS as compared to C-SQS. }

As the temperature increases, however, the SLM distribution becomes more diffuse, spreading probability mass over a wider set of tokens. This expansion of the support set happens selectively, only at the tokens for which the SLM distribution is less sharp. In this setting, C-SQS adaptively expands its support, making it more likely to include the target token and thereby reducing its latency and resampling rate. This conclusion is also reflected in the latency performance. Overall, we observe a clear crossover: $K$-SQS performs better at low temperatures, while C-SQS is more effective at higher temperatures. {In Appendix \ref{ablation_study_appendix} we provide additional experimental results implementing an ablation study on $K$-SQS and C-SQS.}

\textbf{Concluding Remarks:}
In summary, this work demonstrates that efficient compression is critical for overcoming the bandwidth bottleneck in edge–cloud SD. By deriving an information-theoretic characterization of {resampling rate} and introducing the SQS-SD framework, we show how structured sparsification and quantization can effectively reduce communication costs while preserving accuracy. Both $K$-SQS and C-SQS significantly reduce latency, and {resampling rate}, with $K$-SQS proving more effective in low-uncertainty regimes (lower temperatures), while C-SQS achieves the most favorable balance under high-uncertainty regimes (higher temperatures). These results highlight the potential of distribution-aware compression to make edge–cloud LLM inference both practical and scalable.

\textbf{Acknowledgment: }This work was supported by US NSF under Grants CCF-2100013, CNS-2209951, CNS-2317192; by the U.S. Department of Energy, Office of Science, Office of Advanced Scientific Computing under Award DE-SC-ERKJ422; and by NIH under Award R01-CA261457-01A1. The work of O. Simeone was supported by the Open Fellowships of the EPSRC (EP/W024101/1) and by the EPSRC project EP/X011852/1.

    


\bibliography{ref}
\bibliographystyle{plain}
\appendix

\section{Appendix}

\subsection{Sparse Lattice-based Quantization}\label{app:Lattice_Quantization}

Sparse lattice–based quantization (SLQ) \cite{teku2024latency} projects the top-$K$ probability vector onto a structured lattice within the simplex, enabling more efficient representation and finer control over quantization error for a given bit budget. Let us consider a resource-constrained edge device running a SLM that produces a probability distribution $ {q}$ over a vocabulary $V$.  
If $\alpha$ denotes the cumulative probability mass of the least likely tokens, sparse token analysis identifies the subset of most likely tokens $\mathcal{S} \subseteq V$ that accounts for the top $(1-\alpha)$ probability mass. Let $|\mathcal{S}| = K$ be the number of such top tokens.  
The sparse lattice-based quantization technique \cite{teku2024latency} represents this $K$-dimensional probability vector as a point on a discrete lattice within the $K$-dimensional probability simplex  
$\{ {\hat{q}} \in \mathbb{Q}^K | \sum_{i=1}^K  {\hat{q}}[i] = 1\}.$

Given the original SLM distribution $ {q}$, the lattice-quantized distribution over the top-$K$ tokens is denoted $ {\hat{q}}$ and defined as: 
\begin{align*}
\hat{Q}_\ell = \left\{ [ {\hat{q}}[1],  {\hat{q}}[2], \ldots,  {\hat{q}}[K]] \in \mathbb{Q}^K \ \middle|\  {\hat{q}}[i] = \frac{ {b}[i]}{\ell},\ \sum_{i=1}^K  {b}[i] = \ell \right\},
\end{align*}  
where $\ell,  {b}[i]$ are positive integers and $\ell$ is the lattice resolution parameter, and $ {b}[i]$ are integer counts corresponding to each token probability. The overall procedure is shown in Algorithm~\ref{alg:sparse_lattice_quantization}. 
\begin{algorithm}[h]
\caption{Sparse-Lattice–Based Quantization}
\label{alg:sparse_lattice_quantization}
\begin{algorithmic}[1]
\State \textbf{Input:} Prompt $x$, vocabulary $V$, lattice resolution $\ell$, top-$K$ size $K$
\State \textbf{Output:} Sparse-lattice-based quantized probability vector $\hat{ {q}}$

\State \textbf{Probability Vector Generation:} Generate probability distribution $ {q}$ over $V$ from the SLM given prompt $x$
\State \textbf{Top-$K$ Selection:} Identify the subset $\mathcal{S} \subseteq V$ of $K$ tokens with highest probabilities. Extract $ {q} = [q[1], q[2], \dots, q[K]]$.
\State \textbf{Lattice-Based Quantization:}
    \State Compute $b'[i] \gets \lfloor \ell \cdot q[i] + \tfrac{1}{2} \rfloor$, for $i=1,\dots,K$
    \State $\ell' \gets \sum_{i=1}^K b'[i]$
    \If{$\ell' \neq \ell$}
        \State $\zeta[i] \gets b'[i] - \ell \cdot q[i]$, for $i=1,\dots,K$
        \State Sort indices by increasing $\zeta[i]$.
        \If{$\ell' - \ell > 0$}
            \State Decrease the $|\ell' - \ell|$ largest $\zeta[i]$ in $b'[i]$ by $1$
        \Else
            \State Increase the $|\ell' - \ell|$ smallest $\zeta[i]$ in $b'[i]$ by $1$
        \EndIf
    \EndIf
    \State $\hat{ {q}}[i] \gets b'[i] / \ell$, for $i=1,\dots,K$
    \State Compute lexicographic index to represent $ {b} = [b'[1], \dots, b'[K]]$
\State \textbf{Return: } Output $\hat{ {q}}$
\end{algorithmic}
\end{algorithm}

\subsection{Proof of Theorem \ref{the:LASER}}\label{app:proof_theorem1}

\begin{figure}[!ht]
    \centering
    \includegraphics[scale=0.21]{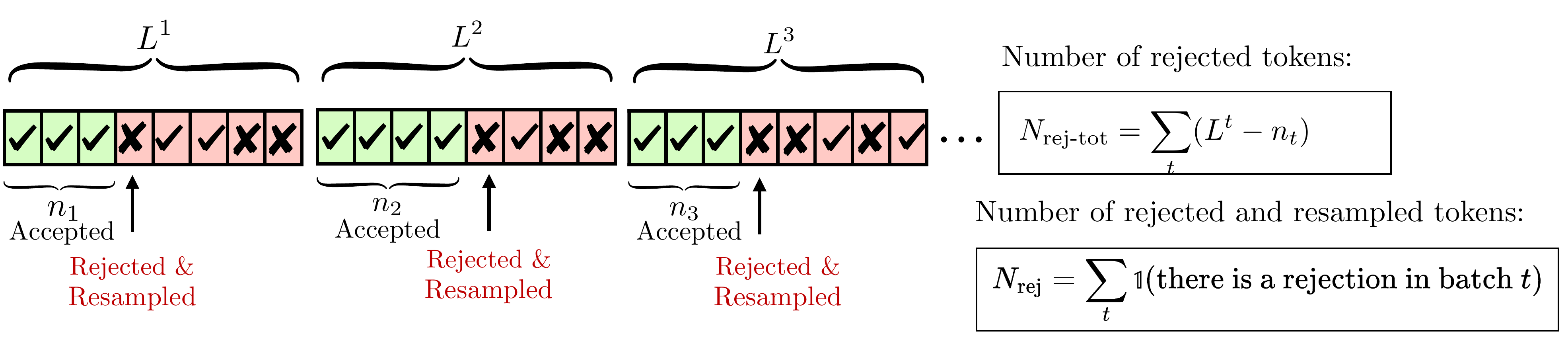}
    \caption{{Illustration of the definition of rejected and resampled tokens, $N_{\textrm{rej}}$,  and of the total number of rejected tokens.}} 
    \label{token_rejections}
\end{figure}

As illustrated in Fig. \ref{token_rejections}, the number of rejected-and-resampled tokens by the LLM, $N_{\text{rej}}$,  
corresponds to the number of times in which the target LLM invalidates a draft proposal and resamples the rejected token. By contrast, the total number of  rejected tokens ($N_{\text{tot-rej}}$ in  Fig. \ref{token_rejections}) corresponds to the total number of SLM generated tokens that were rejected by the target LLM, including all tokens following the resampled token. Computing the expected number of total rejected tokens is non-trivial, since of the rejected tokens do not follow the LLM distribution. For this reason, following  \cite{yin2024theoretical},  we focus on bounding the expected value of rejected and resampled tokens.

  To start, we define a sequence of random variables $R_n \in \{0, 1\}$, 
where $R_n = 1$ indicates that the $n$-th token is rejected. 
The total number of rejections is given by  
\begin{align}
  N_{\mathrm{rej}} = \sum_{n=1}^T R_n.  
\end{align}
Given verified tokens $\{X_t\}^{n-1}_{t=1}$, we compute
\begin{align}
\mathbb{P}(\mathrm{Reject \ at} \ n \mid \{X_t\}^{n-1}_{t=1}).
\end{align}
Let $\tilde{X} \sim \hat q_n(\cdot \mid \{X_t\}^{n-1}_{t=1})$ be the candidate draft token.  
By the law of total probability,
\begin{align}
\mathbb{P}(\mathrm{Reject \ at} \ n \mid \{X_t\}^{n-1}_{t=1})
= \sum_{\tilde{X}} \mathbb{P}(\mathrm{Reject \ at} \ n \mid \tilde{X}, \{X_t\}^{n-1}_{t=1}) 
\, \hat q_n(\tilde{X} \mid \{X_t\}^{n-1}_{t=1}).
\end{align}
By the rejection design of SD, we have 
\begin{align}
\mathbb{P}(\mathrm{Reject \ at} \ n \mid \tilde{X}, \{X_t\}^{n-1}_{t=1})
= 1 - \min\left\{ 1, \frac{p_n(\tilde{X} \mid \{X_t\}^{n-1}_{t=1})}{\hat q_n(\tilde{X} \mid \{X_t\}^{n-1}_{t=1})} \right\}.
\end{align}
Thus, the probability $\mathbb{P}(\mathrm{Reject \ at} \ n \mid \{X_t\}^{n-1}_{t=1})
 $ can be written as 
\begin{align}
\sum_{\tilde{X}} \max\{0, \, \hat q_n(\tilde{X} \mid \{X_t\}^{n-1}_{t=1}) - p_n(\tilde{X} \mid \{X_t\}^{n-1}_{t=1}) \}
= \mathrm{TV}\!\left(\hat q_n(\cdot \mid \{X_t\}^{n-1}_{t=1}), p_n(\cdot \mid \{X_t\}^{n-1}_{t=1})\right).
\end{align}
By the law of total expectation, we arrive at 
\begin{align}
\mathbb{E}[N_{\mathrm{rej}}]
= \sum_{n=1}^T \mathbb{E}[R_n]
= \sum_{n=1}^T \mathbb{E}_{\{X_t\}^{n-1}_{t=1} \sim q}
\left[ \mathrm{TV}\!\left( \hat q_n(\cdot \mid \{X_t\}^{n-1}_{t=1}), \, p_n(\cdot \mid \{X_t\}^{n-1}_{t=1}) \right) \right].
\end{align}

Therefore, 
the expected number of rejections satisfies  
\begin{align}
    \mathbb{E}[N_{\mathrm{rej}}] 
&= \sum_{n=1}^T \mathbb{E}_{\{X_t\}^{n-1}_{t=1} \sim p} 
\left[ \mathrm{TV}\!\left( \hat q_n(\cdot \mid \{X_t\}^{n-1}_{t=1}), \; p_n(\cdot \mid \{X_t\}^{n-1}_{t=1}) \right) \right] \nonumber\\
&\overset{(a)}{\leq} \sum_{n=1}^T \mathbb{E}_{\{X_t\}^{n-1}_{t=1} \sim p} \left[ \mathrm{TV}\!\left(\hat q_n(\cdot \mid \{X_t\}^{n-1}_{t=1}), \; q_n(\cdot \mid \{X_t\}^{n-1}_{t=1}) \right) \right]\nonumber\\
&~~~+\sum_{n=1}^T \mathbb{E}_{\{X_t\}^{n-1}_{t=1} \sim p} 
\left[ \mathrm{TV}\!\left( q_n(\cdot \mid \{X_t\}^{n-1}_{t=1}), \; p_n(\cdot \mid \{X_t\}^{n-1}_{t=1}) \right) \right],
\end{align}
where inequality (a) holds due to the triangle inequality.

    Inspired by \cite{teku2024latency}, we can also bound the sparse lattice-based quantization (SLQ) term $\sum_{n=1}^T \mathbb{E}_{\{X_t\}^{n-1}_{t=1} \sim p} \left[ \mathrm{TV}\!\left(\hat q_n(\cdot \mid \{X_t\}^{n-1}_{t=1}), \; q_n(\cdot \mid \{X_t\}^{n-1}_{t=1}) \right) \right]$. To simply the notation, we denote $ q(\cdot \mid \{X_t\}^{n-1}_{t=1})$ as $q$.  
    
Given the inequality $
\sum_{i \in k} q[i] \ge 1 - \alpha, \quad \alpha \in [0, 1],$ 
we have $\sum_{i \notin k} q[i] \le \alpha.$
Let $S = \sum_{i \in k} q[i].$ 
We define $\bar{q}$ as the normalized probability vector over the $K$ entries, where  
\begin{equation}
\bar{q}[i] = 
\begin{cases}
\frac{q[i]}{S}, & i \in k, \\
0, & \text{otherwise}.
\end{cases}
\end{equation}

We upper bound $\mathrm{TV}(q, \bar{q})$ as follows:
\begin{align}
\mathrm{TV}(q, \bar{q}) 
&= \frac{1}{2} \sum_{i} \left| q[i] - \bar{q}[i] \right| \nonumber \\
&= \frac{1}{2} \left( \sum_{i \in k} \left| q[i] - \frac{q[i]}{S} \right| 
+ \sum_{i \notin k} \left| q[i] - \bar{q}[i] \right| \right) \nonumber\\
&\overset{(a)}{=} \frac{1}{2} \left( \sum_{i \in k} q[i] \left| 1 - \frac{1}{S} \right| 
+ \sum_{i \notin k} q[i] \right) \nonumber\\
&= \frac{1}{2} \left( \sum_{i \in k} q[i] \frac{|S - 1|}{S} 
+ \sum_{i \notin k} q[i] \right)\nonumber \\
&\overset{(b)}{=} \frac{1}{2} \left( \sum_{i \in k} q[i] \frac{1 - S}{S} 
+ \sum_{i \notin k} q[i] \right),
\end{align}
where (a) follows from $\bar{q}[i] = 0$ for all $i \notin k$, and (b) follows from $S \ge 0$.

Following similar steps, we have
\begin{align}
\mathrm{TV}(q, \bar{q}) 
&\overset{(a)}{=} \frac{1}{2} \left( \sum_{i \in k} \frac{q[i]}{S} - \sum_{i \in k} q[i] + \sum_{i \notin k} q[i] \right) = \frac{1}{2} \left( \frac{\sum_{i \in k} q[i]}{S} - \sum_{i \in k} q[i] + \sum_{i \notin k} q[i] \right)\nonumber \\
&\overset{(b)}{=} \frac{1}{2} \left( 1 - \sum_{i \in k} q[i] + \sum_{i \notin k} q[i] \right)\nonumber \\
&\overset{(c)}{=} 1 - \sum_{i \in k} q[i] \overset{(d)}{\le} \alpha,
\end{align}
where (a) follows from $0 \le S \le 1$, (b) from $S = \sum_{i \in k} q[i]$, 
(c) from $\sum_{i \in k} q[i] + \sum_{i \notin k} q[i] = 1$, 
and (d) from $\sum_{i \notin k} q[i] \le \alpha$.

From \cite{teku2024latency}, the distortion due to LQ can be upper bounded as
\begin{equation}
\mathrm{TV}\!\left(\bar{q}, \hat{q}\right) \le \frac{k}{4\ell},
\end{equation}
where $l$ is a positive integer that can be set by users. Therefore $\mathrm{TV}\!\left(m, \hat{q}\right) $ can be upper bound by $  \alpha + \frac{k}{4\ell}$. 
\\\\


\subsection{Proof of Theorem \ref{lem:csqs_bound_all}}
    \label{app:proof_theorem2}

    Recall the threshold update rule introduced in Section ~\ref{sec:conformal} via equation ~\eqref{eq:update}: 
    \begin{equation}
        \beta^t_{n+1}= \beta^t_{n}-\eta \cdot \left(\sum_{x \notin \mathcal{X}_n^t} q_n^t(x)-\alpha\right),
    \end{equation}    
    where $\beta^t_{n+1}$ is the threshold used for token $n$ in batch $t$, $\alpha ^t_n (\mathcal{X}^t_n) = \sum_{x \notin \mathcal{X}_n^{T^t}} q_n^{T^t}(x)$ is the dropped probability mass due to sparsification at step $n$, 
    $\alpha \in (0, 1)$ is the target deviation, and $\eta > 0$ is the learning rate. 
    The pre-batch threshold initialization follows Algorithm \ref{alg:csqs_edge_cloud}. 
    Let the total number of accepted tokens across all batches be $T = \sum\limits_{t}^T L_t$.  
    Throughout this section, the index $t=0,1,\dots,T-1$ enumerates the $T$ accepted tokens in chronological order.

\begin{lemma}\label{lem:TV}
Let $V$ denote the vocabulary, and consider any sparsification strategy.  The total probability mass outside the sparsified token set $\mathcal{X}_n^t$ can be expressed in terms of the total variation (TV) distance between the true distribution $q^{t}$ and the sparsified distribution $\tilde{q}^{t}$:  
\begin{align}
\sum_{x \notin \mathcal{X}^{t}_{n}} q^{t}(x) \;=\; \mathrm{TV}\!\left(q^{t}, \tilde{q}^{t}\right).
\end{align}
\end{lemma}

\begin{proof}
     We start with the definition of $\mathrm{TV}( q^{t}, \tilde{q}^{t})$:
     \begin{align}
         \mathrm{TV}(q^{t}, \tilde{q}^{t}) & = \frac{1}{2} \sum_{x\in\mathbb{V}} |q^{t}(x) -\tilde{q}^{t}(x)|  =\frac{1}{2}\sum_{x\in\mathcal{X}^{t}_{n}}|q^{t}(x) -\frac{q^{t}(x)}{\sum_{x\in\mathcal{X}^{t}_{n}}q^{t}(x)} | + \frac{1}{2}\sum_{x \notin \mathcal{X}^{t}_{n}} q^{t}(x)\nonumber\\
         & = \frac{1}{2}\sum_{x\in\mathcal{X}^{t}_{n}}q^{t}(x) \left( \frac{1}{\sum_{x\in\mathcal{X}^{t}_{n}}q^{t}(x)} - 1 \right)+ \frac{1}{2}\sum_{x \notin \mathcal{X}^{t}_{n}} q^{t}(x)\nonumber\\
         &= \frac{1}{2}\left(1-\sum_{x\in\mathcal{X}^{t}_{n}}q^{t}(x) + \sum_{x \notin \mathcal{X}^{t}_{n}} q^{t}(x)\right) = \sum_{x \notin \mathcal{X}^{t}_{n}} q^{t}(x)
     \end{align}
\end{proof}
This completes the proof of Lemma \ref{lem:TV}. Using this lemma, we can re-write the update rule of \ref{eq:update} as presented in Lemma \ref{lem:telescoping}. We next prove the following result which uses the new alternative form of the threshold update rule. 
\begin{lemma}\label{lem:telescoping}
Given the Lemma \ref{lem:TV} and assume the threshold is updated at each accepted token as
        \begin{equation}
          \beta^{t+1} \;=\; \beta^t \;-\; \eta\,\big(\mathrm{TV}(\tilde q^t, q^t)\;-\;\alpha\big),
          \qquad \eta>0,\;\alpha\in(0,1).
        \end{equation}
        Then the cumulative sparsification distortion satisfies the  identity
        \begin{equation}
          \sum_{t=0}^{T-1} \mathrm{TV}(\tilde q^t, q^t)
          \;=\;
          \alpha\,T \;+\; \frac{|\beta^0 - \beta^{T}|}{\eta}.
        \end{equation}
    \end{lemma}

    This follows via an algebraic telescoping sum. 
    Any asymptotic statement such as
    $\frac{1}{T}\sum_{t=0}^{T-1}\mathrm{TV}(\tilde q^t,q^t)\to\alpha$
    requires a separate argument showing that $| \beta^0-\beta^T |$ is $O(1)$; this will be established by bounding the iterates $\{\beta^t\}$.
    
    \begin{lemma}[Step-size envelope]\label{lem:envelope}
        Since $0\leq \mathrm{TV}(\tilde q_n,q_n)\leq 1$, each update satisfies the inequalities
        \begin{equation}
           -\,\eta(1-\alpha) \;\leq\; \beta_{n+1}-\beta_n \;\leq\; \eta\alpha.
        \end{equation}
      
    \end{lemma}
      Thus, every step moves $\beta$ by at most $\eta$ in either direction.
    \begin{lemma}[Universal bound on $\beta$]\label{lem:universal}
        The sequence $(\beta_n)$ is uniformly bounded:
        \begin{equation}
           -\,\eta(1-\alpha) \;\leq\; \beta_n \;\leq\; 1+\eta\alpha,
           \qquad \forall n\geq 0.
        \end{equation}
    \end{lemma}
    
    \begin{proof}
        If $\beta_n<0$, then thresholding keeps the full support and hence $\mathrm{TV}(\tilde q_n,q_n)=0$. 
        The update becomes
        $\beta_{n+1}=\beta_n+\eta\alpha > \beta_n$, so $\beta$ is forced upward.
        By Lemma~\ref{lem:envelope}, the largest one-step overshoot below 0 is $-\eta(1-\alpha)$.
        
        If $\beta_n>1$, then thresholding discards all but the top outcome,
        so $\mathrm{TV}(\tilde q_n,q_n)=1$. 
        Since $\alpha<1$, we obtain $\beta_{n+1}=\beta_n-\eta(1-\alpha)<\beta_n$, so $\beta$ is forced downward.
        By Lemma~\ref{lem:envelope}, the largest one-step overshoot is $ 1 +\eta\alpha$.
        Combining the two cases yields the stated bound.
    \end{proof}

        Plugging Lemma~\ref{lem:universal} into Lemma~\ref{lem:telescoping} gives
        \begin{equation}
           \sum_{n=1}^{T}\mathrm{TV}(\tilde q_n,q_n)
           \;\leq\; \alpha T \;+\; \frac{|\beta_0| + 1 + \eta}{\eta},
        \end{equation}
        and hence
        \begin{equation}
           \frac{1}{T}\sum_{n=1}^{T}\mathrm{TV}(\tilde q_n,q_n)
           \;\leq\; \alpha \;+\; O(T^{-1}), 
        \end{equation} as desired. Furthermore,      if $\eta = c\,T^{-1/2}$ and $\alpha = d\,T^{-1/2}$ with $c,d>0$, then
        \begin{equation}
            \sum_{n=1}^{T}\mathrm{TV}(\tilde q_n,q_n)
            = \mathcal{O}(\sqrt{T}),
            \qquad
            \frac{1}{T}\sum_{n=1}^{T}\mathrm{TV}(\tilde q_n,q_n)
            = \alpha + \mathcal{O}(T^{-1/2}).
        \end{equation}
        This completes the proof.
\subsection{Additional Experimental Results \label{ablation_study_appendix}} { In this part of the Appendix, we present an ablation study that investigates the effects of the parameters $K$ and $\beta$ for the $K$-SQS and C-SQS methods, respectively, while varying the temperature $T$. We also report results demonstrating that C-SQS without adaptivity (i.e., learning rate $\eta = 0$) exhibits higher latency and re-sampling rates compared to the adaptive version ($\eta > 0$), thereby highlighting the benefits of adaptive parameter updates for $\beta$.
}
\subsubsection{Impact of Hyperparameters $K$ and $\beta$} 
\begin{figure*}[h]
    \centering
    \begin{subfigure}[t]{0.5\textwidth}
        \centering
        \includegraphics[width=\linewidth]{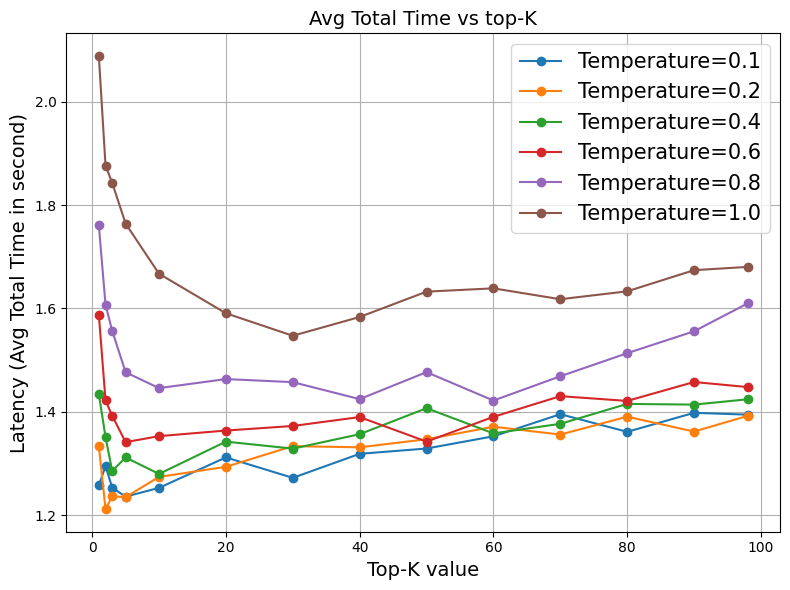}
        \label{fig:latency_ksqs}
    \end{subfigure}
    \hfill
    \begin{subfigure}[t]{0.48\textwidth}
        \centering
        \includegraphics[width=\linewidth]{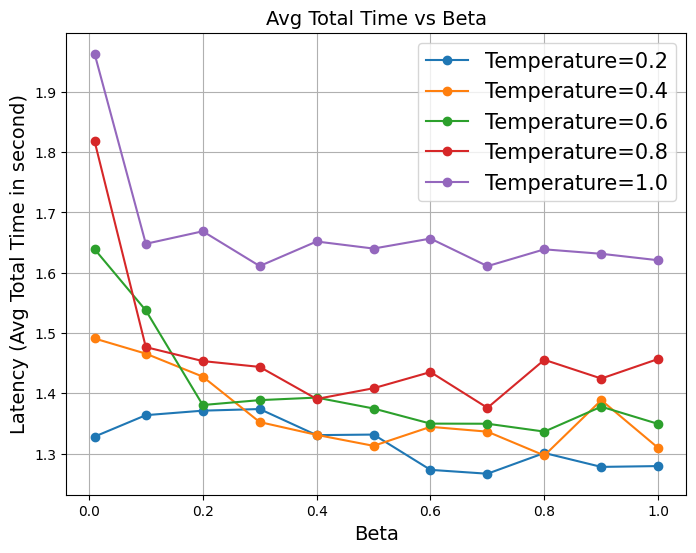}
        \label{fig:latency_csqs}
    \end{subfigure}
    \caption{{Latency  for $K$-SQS and C-SQS methods versys $K$ and $\beta$, respectively, across varying temperature settings. }}
    \label{fig:latency_comparison}
\end{figure*}

{First we have analyzed the effect of the parameters $K$ and $\beta$ over varying temperature for $K$-SQS and C-SQS methods. Figure \ref{fig:latency_comparison} shows the latency performance for $K$-SQS and C-SQS  across varying temperature settings. Both schemes demonstrate consistent performance trends, with latency generally increasing as the temperature and consequently the sampling uncertainty-rises. 

For $K$-SQS, performance is highly dependent on the choice of the $K$-value: smaller $K$ values yield lower latency but may reduce stability, while larger $K$ values improve robustness at the cost of increased computation. In contrast, C-SQS leverages adaptive threshold tuning through the parameter $\beta$, enabling dynamic control over resampling and thereby achieving smoother latency–accuracy trade-offs. 

Collectively, these findings indicate that $K$-SQS and C-SQS are complementary: $K$-SQS performs optimally in low-uncertainty regimes with well-chosen $K$, whereas C-SQS excels under higher-uncertainty conditions due to its adaptive thresholding mechanism.}
\newpage
\subsubsection{Benefits of Adaptivity in C-SQS}
\begin{figure}[h]
    \centering
    \includegraphics[scale = 0.21]{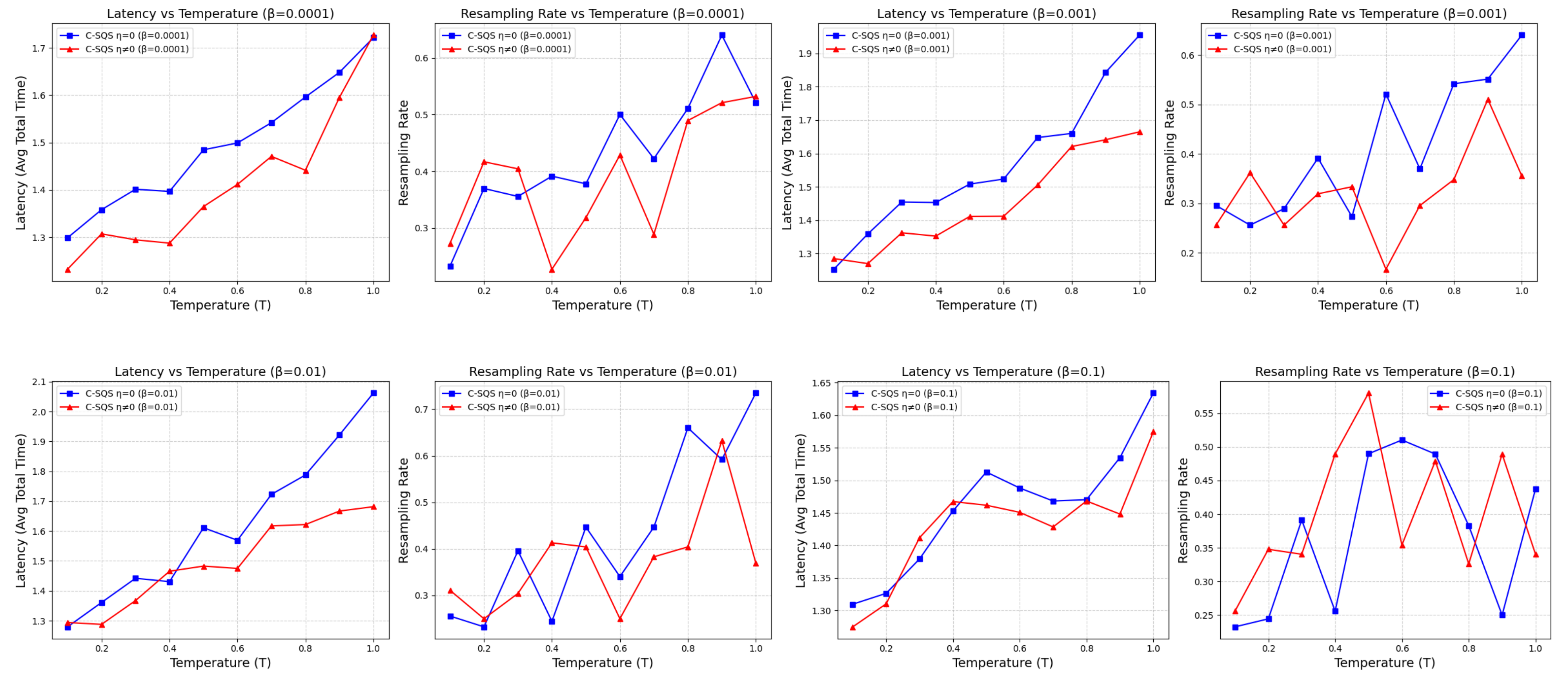}
    \caption{{Latency and resampling rate as a function of temperature  for C-SQS with and without adaptivity. }}
    \label{fig:csqs_adaptivity}
\end{figure}

{Figure~\ref{fig:csqs_adaptivity} presents the relationship between temperature and (a) latency, and (b) resampling rate, across different initial threshold ($\beta$) values, comparing adaptive ($\eta > 0$) and non-adaptive ($\eta = 0$) configurations of C-SQS. It is observed that incorporating adaptivity during $\beta$-updates significantly improves efficiency. Notably, for smaller $\beta$ values that represent more conservative acceptance thresholds, the adaptive variant consistently yields lower latency and reduced resampling rates relative to the non-adaptive baseline. This indicates that adapting the threshold $\beta$ enables C-SQS to better regulate resampling behavior, effectively balancing stability and responsiveness under varying uncertainty (temperature) conditions.}

\subsubsection{$K$-SQS vs C-SQS}
\begin{figure}
    \centering
    \includegraphics[scale = 0.21]{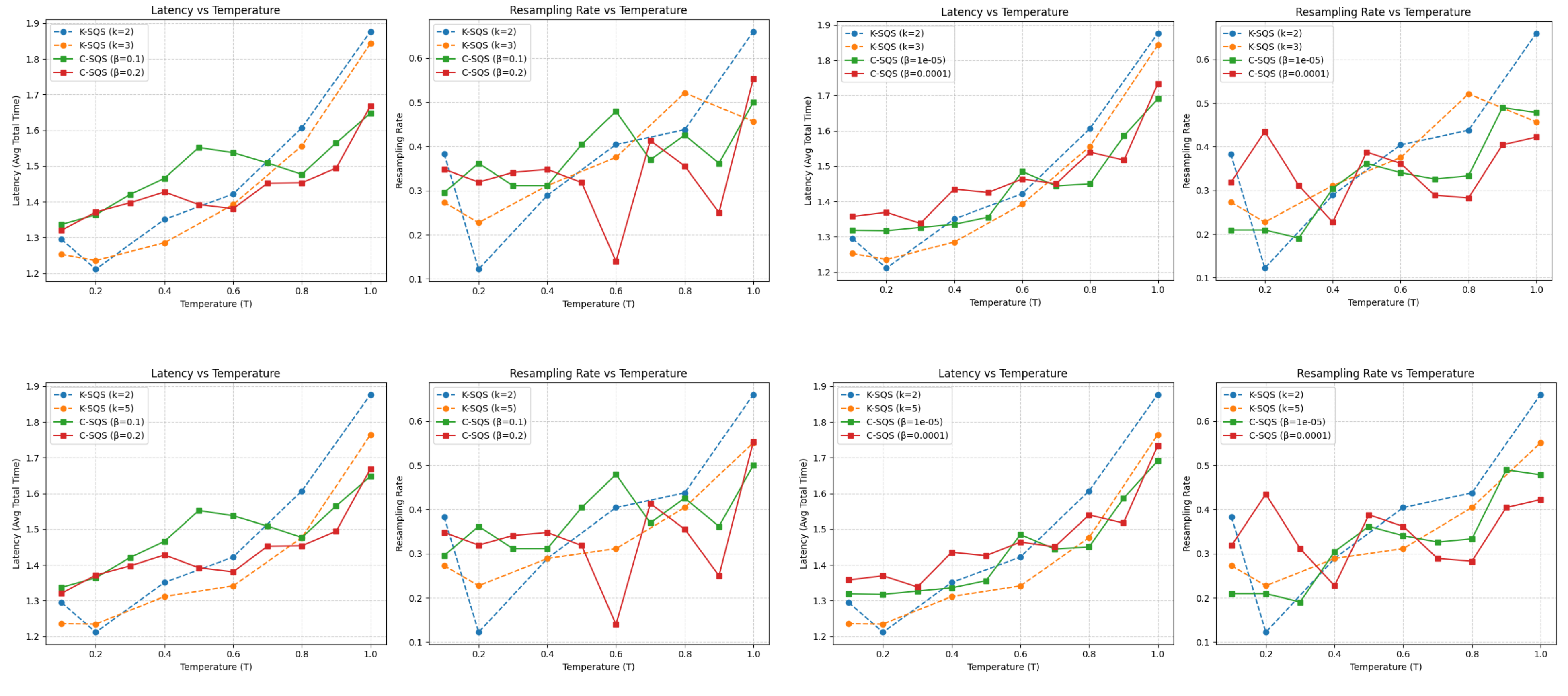}
    \caption{{Latency and resampling rate of $K$-SQS and C-SQS across varying temperature ($T$) settings.
    }}
    \label{fig:ksqs_vs_csqs_appendix}
\end{figure}
{In Figure~\ref{fig:ksqs_vs_csqs_appendix}, we compare the performance of $K$-SQS and C-SQS under varying temperature ($T$) values, illustrating their latency and resampling characteristics. 
As temperature increases, indicating higher sampling uncertainty, both methods show a general rise in latency and resampling rate due to more frequent token rejections. 
For $K$-SQS, performance is highly influenced by the selection of the $K$-value: smaller $K$ values yield faster but less stable performance, while larger $K$ values improve reliability at the cost of increased latency. 

In contrast, C-SQS leverages its adaptive threshold $\beta$, allowing it to dynamically adjust to temperature changes and maintain a more balanced latency, accuracy trade-off. 
Overall, these results highlight that $K$-SQS is preferable in low-uncertainty regimes (lower temperature regimes), whereas C-SQS demonstrates greater robustness and efficiency in higher-uncertainty conditions due to its adaptive mechanism.}

\end{document}